\pdfoutput=1
\documentclass{article}
\PassOptionsToPackage{numbers,sort&compress}{natbib}
\usepackage[final,dblblindworkshop]{neurips_2026}
\workshoptitle{Interpreting Agent Behavior (IAB)}
\usepackage[T1]{fontenc}
\usepackage{amsmath}
\usepackage{amssymb}
\usepackage{booktabs}
\usepackage{graphicx}
\usepackage{array}
\usepackage[expansion=false]{microtype}
\usepackage{xcolor}
\usepackage{hyperref}
\usepackage{enumitem}
\setlist{itemsep=0pt,topsep=0pt,parsep=0pt,leftmargin=*}
\hypersetup{colorlinks=true,linkcolor=blue!55!black,citecolor=blue!55!black,urlcolor=blue!55!black,pdfauthor={Egor Pakhomov, Erik Nijkamp}}
\title{Attributing Agent Failures to Their History Needs an Executed Counterfactual: Toward a Reporting Standard}
\author{%
  Egor Pakhomov \\
  Salesforce AI Research \\
  \texttt{epakhomov@salesforce.com} \\
  \And
  Erik Nijkamp \\
  Salesforce AI Research \\
  \texttt{erik.nijkamp@salesforce.com} \\
}

\begin{document}
\maketitle

\begin{abstract}
Published explanations of why long-horizon agents fail are commonly taxonomies labeled over completed trajectories: a judge reads a transcript and names the failure. We argue that a claim attributing an agent's behavior to its history should carry intervention-verified evidence, per instance or via a validated estimator: freeze an executable mid-trajectory state, alter only the history the agent sees, and continue---executed rather than inferred from the transcript, at a state whose restoration has been measured, with a replay certificate matched to the estimand. Each of the four manipulation classes we name has a precedent, and several methods already execute interventions from checkpoints, re-executed prefixes or recorded conversation states (for example DoVer, REFLECT, Min et al., Who\&When Pro, Causal Agent Replay), but none reports a measured replay certificate. Two cautionary measurements, neither applying the standard, follow: a carrier taxonomy whose ``concentration'' could not be separated from benchmark construction
(adjusted odds ratio 1.15, interval crossing 1),
and hash-identical snapshots that, restored into identically seeded processes, agreed with one another in 24 of 24 states while differing from the live continuation in 9 and 20 of 24. Restoring the captured position of a random stream made all 24 agree with the live run.

\end{abstract}

\section{The correlational trap}\label{sec:trap}

A common way of explaining agent failure is to collect finished trajectories, have a human or LLM judge read them, and assign each a taxonomy label~\citep{song2025aegis,gao2026interpret}.

HORIZON~\citep{wang2026horizon} attributes failures across 3,100+ trajectories via a 7-category taxonomy applied by an LLM judge
and never resumes from an intermediate state, so its ``History Error Accumulation'' label is post-hoc.
MAST~\citep{cemri2025mast} derives 14 failure modes from 150 traces and reports that verification failures ``appear frequently even in successful runs.''
Both intervene only at task or system level (horizon extension; design changes worth $+9.4\%$ and $+15.6\%$); the failure \emph{labels} remain post-hoc.

AgentLocate~\citep{xia2026agentlocate} formalizes the decisive step counterfactually, as the earliest step whose single correction flips the outcome (Who\&When's definition~\citep{zhang2025whowhen}), and then states that its judge ``does not execute'' the counterfactual but ``approximates this criterion by reasoning over the logged trajectory,'' as RAFFLES~\citep{zhu2025raffles} and DuoTrace~\citep{zhang2026duotrace} also do offline.
Who\&When Pro~\citep{liu2026whowhenpro} does execute: 12,326 golden labels from what it describes as exact prefix replay with one injected error.
Who\&When Pro argues that regenerating the whole trajectory, as AgenTracer~
\citep{zhang2025agentracer} and the Aegis attribution pipeline~\citep{kong2025aegis} do, lets the pre-injection portion drift (no rate is reported).
The field already \emph{defines} causes counterfactually; AgentLocate and the Who\&When evaluators still attribute from transcripts, and the executing methods of \S\ref{sec:standard} lack a measured replay certificate. In a two-coder LLM census of the 41 failure-analysis papers in our reading list (Appendix~\ref{app:census}), the coders disagreed on how many manipulate history or state mid-run (coder 1: 22; coder 2: 13, under a stricter boundary). Both found few that restore an executable intermediate state and continue the same agent: one and three respectively, and zero and two once Min et al.\ is treated as prefix re-execution. Two to three papers' reading notes were too thin to code; secondhand notes give no upper bound.

A label assigned by reading a completed transcript, blinded or not, cannot distinguish a \emph{cause} of failure from something that \emph{co-occurs} with it.

\paragraph{Position.} A claim that an agent behaved as it did \emph{because of} something in its history should carry a \textbf{history-manipulation control at a frozen executable state}: the same agent, continued from the same restored state, under a history that differs in exactly the respect the explanation names. Causal interpretability already argues that a representational claim needs an intervention on a fixed forward pass~\citep{vig2020causal,geiger2021causal}; the frozen executable state plays that role for claims about history and state. We name four manipulation classes and their precedents (\S\ref{sec:standard}), report two cautionary measurements (\S\ref{sec:measurements}), and propose an agenda (\S\ref{sec:agenda}).

\section{History manipulation at frozen executable states}\label{sec:standard}

\paragraph{Setup.} Let a trajectory reach step $t$ with environment state $s_t$, process state $p_t$ (random streams, open handles; caches derived from the history are recomputed from the manipulated history $h'_t$ defined below), executable state $x_t = (s_t, p_t)$, and agent-visible history $h_t$. Three properties are distinct: (i) the fork \emph{restores} $x_t$ (now cheap with sandbox checkpointing~\citep{deltabox2026}; certified by a \emph{replay certificate}: a measured restored-vs-reference agreement rate under a fixed probe policy); (ii) \emph{exact replay}: continuations under a deterministic probe policy and coupled exogenous randomness are bitwise identical; (iii) continuations under independent randomness are \emph{distributionally} equal. A \emph{frozen executable state} is one where (i) has been measured, not assumed, along the probe run and meets a rate or margin fixed in advance for the estimand. Pathwise estimands additionally require (ii); distributional ones require replicated continuations against an estimand-matched control; the certificate is probe-specific, and \S\ref{sec:rng} measures (i) and (ii).
A \textbf{history manipulation} replaces $h_t$ by $h'_t$ while holding the frozen state fixed, then continues under the same policy and compares arms. Four classes organize the history-based explanations in the precedents below:
\begin{enumerate}
\item \textbf{Healed history}: errors repaired in place or an oracle history substituted.
\item \textbf{Provenance swap}: byte-identical content re-attributed to a different source role (self vs.\ prior agent, tool, user).
\item \textbf{Placebo / length-matched synthetic history}: same token and turn count, inert or idealized content.
\item \textbf{Counterfactual injection}: a controlled error, stale value, or compaction summary inserted where none occurred.
\end{enumerate}
Holding $s_t$ fixed while editing $h_t$ targets a \emph{controlled direct effect} of what the agent is shown, given the realized world; if the error also changed the world, a healed history paired with the damaged state is off-support; the total effect needs a fork before the event or a joint history-and-state arm.

\paragraph{Precedents.} Two precedents re-execute a recorded prefix to rebuild the state (Min et al., recipe hash; Who\&When Pro, cached static observations plus an abort-on-exception guard); Causal Agent Replay (CAR) holds the recorded actions before a step fixed and re-decides from it; REFLECT resumes from a recorded conversation state; AgentDebug~\citep{zhu2025agentdebug} re-rolls out from the critical step with feedback, mechanism unstated; DoVer~\citep{ma2025dover} edits plans or messages at a checkpointed step and resumes; CausalFlow~\citep{bonagiri2026causalflow} replaces a step and re-executes downstream where an executor exists (predicting it otherwise); TraceElephant~\citep{chen2026traceelephant} replays to the identified step, alters its input and observes three further steps.

None reports the measured replay-agreement certificate its estimand requires, pathwise or distributional.
Forking both arms from one restored instance validates their contrast; the live-vs-restored control ties that instance to the observed trajectory; separately restored arms need a between-arm identity check.
\begin{itemize}
\item \emph{Healed history and controlled-error injection} (no executable environment).
\citet{sinha2025illusion} replay the turn-100 step under injected histories with controlled error rates;
in memory-augmented chat, \citet{xiao2026errorsmemories} repair the question, the memory, or both and continue.
\item \emph{Length-matched synthetic history} (repeated social dilemmas).
\citet{liu2026memorycurse} replace $80-X$ of 80 rounds with synthetic cooperative records:
restoring cooperation from 6.9\% to 97.4\% at $X = 2$ for one model (less for weaker ones).
\item \emph{Provenance swap} (single claims in math/logic traces).
\citet{chen2026selfcorrection} move byte-identical wrong claims between the model's own thought and user/tool/memory roles; explicit-correction rates rise by 23--93 points.
\item \emph{State-side counterfactual} (environment, not history).
ChainSWE~\citep{jin2026chainswe} contrasts oracle-reset against inherited repositories: 58.9\% $\rightarrow$ 36.5\% per bug, 48\% of downstream failures attributable to inherited state.
\item \emph{Step-level counterfactuals by prefix hold or conversation-state resume} (attribution). Causal Agent Replay~\citep{shah2026causalreplay} applies do-operations at a step (resample, action, observation, history or policy), holds the recorded actions before the step fixed and re-runs forward to an outcome distribution with intervals, but validates only on two synthetic causal models and reports no numeric action-match rate or live-vs-restored baseline.
REFLECT~\citep{lin2026reflect} injects a repair message into a recorded conversation state and re-runs the agent at temperature 0 (placebo and wrong-step controls at temperature 0.2), but reports no replay-fidelity measurement and does not state how the rollback state is restored.
\item \emph{Compaction summary vs.\ raw history by recipe-hash-checked prefix re-execution} (class 4; closest executable-surface precedent; no measured replay certificate).
\citet{min2026trace} re-create what their protocol treats as the same AppWorld~\citep{trivedi2024appworld} state at each of 590 compaction boundaries by re-executing the recorded prefix under a hash of the recipe, not of the reached state (whether the states matched was not measured; Appendix~\ref{app:trace}), and continue under two arms, a less-compacted control and the new summary, three to five samples each,
released as TRACE~\citep{trace2026repo}.
\end{itemize}
We propose transplanting interventionist evidence and replay validation, established in causal inference and record-and-replay, to agent-history attribution. Among the empirical LLM-agent papers we examined, Causal Agent Replay comes closest to the reporting \emph{norm}; the increment here is narrow: a reported numeric replay certificate, an unedited live-vs-restored control, and an estimand-matched certificate per regime, the first of which CAR names but does not quantify. Resampling over reasoning traces~\citep{bogdan2025anchors} shares the estimand without an environment to restore; explainable-RL methods intervene, when they do, on actions and policies, not history~\citep{zhang2023understanding,kim2025talktoagent}.
 Our census excluded position papers without coded failure data, so it cannot show the norm is unproposed.

\section{Two cautionary measurements}\label{sec:measurements}

Neither applies the standard; amendments are disclosed (\S\ref{sec:discussion}, Appendices~\ref{app:carrier}--\ref{app:trace}).

\subsection{A carrier taxonomy confounded with benchmark construction}\label{sec:carrier}

We asked an observational question of ActBench~\citep{yao2026actbench}, an agent-safety corpus (300 task pairs, 24,000 trajectories):
which \emph{carrier} delivers the payloads that convert into unsafe actions? Deterministic carrier labels covered 266/300 task pairs, a same-agent LLM consistency check over raw bundles (rule labels visible) agreed on 50/50 sampled cases,
and a degeneracy guard, declared after the audit tallies but before any outcome statistic, held that if fewer than three classes survived the pre-specified collapse, the concentration and ranking layers were \emph{not evaluable}.
The guard fired: \texttt{service\_response} (225 task pairs) vs.\ \texttt{local\_file\_state} (41) remained.
The headline statistic, 84.9\% [80.1, 89.4] of 9,199 attack-positive trajectories (ActBench's attack-pass label, AGS $\geq 0.8$) carried by service responses%
, tracks task-pair allocation by construction ($225/266 = 84.6\%$ of labeled task pairs are service-carried);
the behavior- and attack-method-adjusted odds ratio was 1.148 [0.627, 2.016], crossing 1;
and first-exposure recovery was 96.65\% for service responses against 69.31\% for the collapsed local-file/memory/tool-metadata class, so even exposure-conditional conversion rates are confounded by \emph{measurement recoverability} (Appendix~\ref{app:carrier}).
The measurement failed its pre-outcome amended per-class recovery floor ($69.31\% < 70\%$).
The share is a valid descriptive statistic for the 266 labeled task pairs; these data cannot separate benchmark construction from agent response and identify no service-versus-local carrier effect in either direction.

\subsection{Hash-identical snapshots need not replay exactly}\label{sec:rng}

We tested whether hash-identical states replay exactly with our own snapshot layer over AgentCE-Bench~\citep{yang2026agentce}, with a fixed ground-truth continuation policy and no model in the loop, on 24 factorial-cell states.
A five-clone state-hash certificate passed every state tried with the stochastic tool-failure channel off (144 synthetic verdicts over 108 distinct states, plus 17 of 17 model-generated),
and an in-snapshot sensitivity check flagged 45/45 single-field edits (blind to omitted state by construction).
With the channel on, five unseeded naive restorations of each hash-identical state disagreed among themselves in 17/24 states at failure rate 0.05 and 22/24 at 0.20.
The tool-failure draw consumes the process-global random stream, which the upstream runner seeds at start-up and re-seeds while building its task cache, and no snapshot field records the stream's position.

In a seeded replication, two fresh single-threaded re-executions agreed in 24/24 states, as did a state rebuilt by replaying the recorded prefix. Under one seed per state, byte-identical snapshots restored into processes seeded with that state's seed diverged from the live run in 9/24 and 20/24 states (0.38 and 0.83; independent-sequence reference 0.45 and 0.84, Appendix~\ref{app:replay}); under the runner's single fixed seed, in 18/24 and 24/24. In both regimes the five clones agreed with one another in 24/24. A clone-vs-clone gate whose clones share a seed position passes and is still wrong; a separately executed reference continuation, live or prefix-rebuilt, detects it.
Capturing the stream state alongside the snapshot and restoring it before each continuation removed every divergence in the restoration arms: live-vs-restored 0/24 at both rates in the unseeded experiment and in the seeded replication, and five-clone 0/24 at both rates in the unseeded experiment.
The snapshot omitted the stream position that \S\ref{sec:standard} counts as process state. Capturing it yields exact replay for this probe by coupling the exogenous randomness across clones (common random numbers~\citep{law2015simulation}; record-and-replay~\citep{ocallahan2017rr}); it does not show the executable state is complete.

Two consequences follow. First, final-grader divergence under the oracle retry policy was 0/24 in every condition: an outcome-only metric would have hidden the problem, and whether realization-level infidelity moves model-driven behavior is unmeasured.
Second, a replay-based interpretation should state whether it needs exact replay or distributional equality and show the matching \emph{measured} agreement rate against an unedited-history live-vs-restored control.

\section{Discussion and limitations}\label{sec:discussion}

\paragraph{Claims.} The four classes have precedents (\S\ref{sec:standard}); outcome-labeled taxonomies cannot by themselves establish causal attribution to history.
Neither measurement establishes a behavioral result: Measurement~3.1 was declared non-evaluable (\S\ref{sec:carrier}) and Measurement~3.2 is model-free.

\paragraph{Not measured.} A third measurement, a prefix taxonomy over the TRACE boundaries, is not reported here: on artifact review its naive has-written label proved a trajectory-phase proxy, so it does not measure its construct (Appendix~\ref{app:trace}).

\paragraph{Amendments.}
In reproducing Min et al.'s released scan, a byte-identity criterion failed on two bootstrap-bearing tables under an unsupported interpreter and, on a rationale later found mistaken,
was amended to identical point estimates plus overlapping intervals; a supported interpreter reproduced all six tables byte-identically (Appendix~\ref{app:trace}). The seeded replication's pre-registration was frozen before execution; two arms were added by dated amendment after the fixed-seed results were seen, and the per-state arms were designated primary after both result sets were known (Appendix~\ref{app:replay}).

\section{Agenda}\label{sec:agenda}

\begin{enumerate}
\item \textbf{The counterfactual norm.} A claim that ``the agent failed because of $X$ in its history'' should carry per-instance intervention evidence (continuations from a frozen state with $X$ removed, replaced, or re-attributed; replicated unless replay is pathwise-certified) or an estimator validated on a representative, intervention-verified test set with reported calibration and uncertainty; otherwise the attribution is not intervention-verified under this standard.
\item \textbf{Replay certificates before interpretation.} Report the certificate the estimand needs, on the surface and provider used: coupled-randomness agreement (\S\ref{sec:rng}) against an unedited-history control (pathwise), or restored-to-live equivalence within a prespecified margin on a named metric, with replicated controls (distributional), plus a detector-power control. For hosted models the pathwise route may be unavailable~\citep{atil2024nondeterminism,yuan2025numerical,shah2026causalreplay}, as on the one path we tried (Appendix~\ref{app:replay});
a paired-contrast design is in Appendix~\ref{app:factorial} (proposed, not run).
\end{enumerate}

\begin{ack}
This work was funded by Salesforce.
\end{ack}

\bibliographystyle{plainnat}
\bibliography{main}

\begin{thebibliography}{39}
\providecommand{\natexlab}[1]{#1}
\providecommand{\url}[1]{\texttt{#1}}
\expandafter\ifx\csname urlstyle\endcsname\relax
  \providecommand{\doi}[1]{doi: #1}\else
  \providecommand{\doi}{doi: \begingroup \urlstyle{rm}\Url}\fi

\bibitem[Atil et~al.(2024)Atil, Aykent, Chittams, Fu, Passonneau, Radcliffe,
  Rajagopal, Sloan, Tudrej, Ture, Wu, Xu, and Baldwin]{atil2024nondeterminism}
Berk Atil, Sarp Aykent, Alexa Chittams, Lisheng Fu, Rebecca~J. Passonneau, Evan
  Radcliffe, Guru~Rajan Rajagopal, Adam Sloan, Tomasz Tudrej, Ferhan Ture, Zhe
  Wu, Lixinyu Xu, and Breck Baldwin.
\newblock {Non-Determinism of "Deterministic" LLM Settings}.
\newblock \emph{arXiv preprint arXiv:2408.04667}, 2024.

\bibitem[Barres et~al.(2025)Barres, Dong, Ray, Si, and
  Narasimhan]{barres2025tau2}
Victor Barres, Honghua Dong, Soham Ray, Xujie Si, and Karthik Narasimhan.
\newblock {\mbox{$\tau^2$-Bench}: Evaluating Conversational Agents in a
  Dual-Control Environment}.
\newblock \emph{arXiv preprint arXiv:2506.07982}, 2025.

\bibitem[Bogdan et~al.(2025)Bogdan, Macar, Nanda, and Conmy]{bogdan2025anchors}
Paul~C. Bogdan, Uzay Macar, Neel Nanda, and Arthur Conmy.
\newblock {Thought Anchors: Which LLM Reasoning Steps Matter?}
\newblock \emph{arXiv preprint arXiv:2506.19143}, 2025.

\bibitem[Bonagiri et~al.(2026)Bonagiri, Borkar, Anderias, Rafatirad, and
  Homayoun]{bonagiri2026causalflow}
Akash Bonagiri, Devang Borkar, Gerard~Janno Anderias, Setareh Rafatirad, and
  Houman Homayoun.
\newblock {CausalFlow: Causal Attribution and Counterfactual Repair for LLM
  Agent Failures}.
\newblock \emph{arXiv preprint arXiv:2605.25338}, 2026.

\bibitem[Cemri et~al.(2025)Cemri, Pan, Yang, Agrawal, Chopra, Tiwari, Keutzer,
  Parameswaran, Klein, Ramchandran, Zaharia, Gonzalez, and
  Stoica]{cemri2025mast}
Mert Cemri, Melissa~Z. Pan, Shuyi Yang, Lakshya~A. Agrawal, Bhavya Chopra,
  Rishabh Tiwari, Kurt Keutzer, Aditya Parameswaran, Dan Klein, Kannan
  Ramchandran, Matei Zaharia, Joseph~E. Gonzalez, and Ion Stoica.
\newblock {Why Do Multi-Agent LLM Systems Fail?}
\newblock \emph{arXiv preprint arXiv:2503.13657}, 2025.

\bibitem[Chen et~al.(2026{\natexlab{a}})Chen, Su, Lin, Li, and
  Chiang]{chen2026selfcorrection}
Kuan-Yen Chen, Fang-Yi Su, Shih-Yen Lin, Bao Li, and Jung-Hsien Chiang.
\newblock {The Self-Correction Illusion: Role Relabeling Gates Explicit Error
  Flagging in Large Language Models}.
\newblock \emph{arXiv preprint arXiv:2606.05976}, 2026{\natexlab{a}}.

\bibitem[Chen et~al.(2026{\natexlab{b}})Chen, Wang, Mu, Wang, Liu, Feng, and
  Wang]{chen2026traceelephant}
Mengzhuo Chen, Junjie Wang, Fangwen Mu, Yawen Wang, Zhe Liu, Huanxiang Feng,
  and Qing Wang.
\newblock {Seeing the Whole Elephant: A Benchmark for Failure Attribution in
  LLM-based Multi-Agent Systems}.
\newblock \emph{arXiv preprint arXiv:2604.22708}, 2026{\natexlab{b}}.

\bibitem[Clopper and Pearson(1934)]{clopper1934}
C.~J. Clopper and E.~S. Pearson.
\newblock {The Use of Confidence or Fiducial Limits Illustrated in the Case of
  the Binomial}.
\newblock \emph{Biometrika}, 26\penalty0 (4):\penalty0 404--413, 1934.

\bibitem[Dong et~al.(2026)Dong, He, Liu, Hou, Du, Xu, Yu, Yang, Xia, and
  Chen]{deltabox2026}
Yunpeng Dong, Jingkai He, Shiqi Liu, Yuze Hou, Dong Du, Zhonghu Xu, Si~Yu,
  Baochuan Yang, Yubin Xia, and Haibo Chen.
\newblock {DeltaBox: Scaling Stateful AI Agents with Millisecond-Level Sandbox
  Checkpoint/Rollback}.
\newblock \emph{arXiv preprint arXiv:2605.22781}, 2026.

\bibitem[Gao et~al.(2026)Gao, Sun, Huang, Koevering, Ji, Huang, Shi, Lu, Xiao,
  Khashabi, and Dredze]{gao2026interpret}
Jie Gao, Kaiser Sun, {Jen-tse} Huang, Katherine~Van Koevering, Sijie Ji, Heyuan
  Huang, Weiyan Shi, Zhuoran Lu, Ziang Xiao, Daniel Khashabi, and Mark Dredze.
\newblock {How to Interpret Agent Behavior}.
\newblock \emph{arXiv preprint arXiv:2605.13625}, 2026.

\bibitem[Geiger et~al.(2021)Geiger, Lu, Icard, and Potts]{geiger2021causal}
Atticus Geiger, Hanson Lu, Thomas Icard, and Christopher Potts.
\newblock {Causal Abstractions of Neural Networks}.
\newblock \emph{Advances in Neural Information Processing Systems 34 (NeurIPS
  2021); arXiv:2106.02997}, 2021.

\bibitem[Jin et~al.(2026)Jin, Tung, Li, Shi, She, Jia, Zhao, Xia, Du, Pei,
  Zhang, Qi, Duan, Lee, and Jin]{jin2026chainswe}
Qirui Jin, Lingching Tung, Kenan Li, Qiyang Shi, Yushi She, Huanzhong Jia,
  Harrison Zhao, Kejing Xia, Zhenbang Du, Jiaxin Pei, Zhenyu Zhang, Zhen Qi,
  Yuyan Duan, Wenke Lee, and Zijian Jin.
\newblock {ChainSWE: Benchmarking Coding Agents on Multi-Bug Software
  Maintenance}.
\newblock \emph{arXiv preprint arXiv:2607.02606}, 2026.

\bibitem[Kim et~al.(2025)Kim, Chen, Li, and Lee]{kim2025talktoagent}
Haechang Kim, Hao Chen, Can Li, and Jong~Min Lee.
\newblock {TalkToAgent: A Human-centric Explanation of Reinforcement Learning
  Agents with Large Language Models}.
\newblock \emph{arXiv preprint arXiv:2509.04809}, 2025.

\bibitem[Kong et~al.(2025)Kong, Zhang, Yin, Zhang, Zhang, Chen, Zhang, Zhang,
  Zhu, and Feng]{kong2025aegis}
Fanqi Kong, Ruijie Zhang, Huaxiao Yin, Guibin Zhang, Xiaofei Zhang, Ziang Chen,
  Zhaowei Zhang, Xiaoyuan Zhang, Song-Chun Zhu, and Xue Feng.
\newblock {Aegis: Automated Error Generation and Attribution for Multi-Agent
  Systems}.
\newblock \emph{arXiv preprint arXiv:2509.14295}, 2025.

\bibitem[Law(2015)]{law2015simulation}
Averill~M. Law.
\newblock \emph{{Simulation Modeling and Analysis}}.
\newblock McGraw-Hill, New York, 5th edition, 2015.

\bibitem[Lin et~al.(2026)Lin, Wang, Kwok, Guo, Nale, Fleming, and
  Cheng]{lin2026reflect}
Xiaofeng Lin, Yingxu Wang, Tung Sum~Thomas Kwok, Daniel Guo, Sahil~Arun Nale,
  Charles Fleming, and Guang Cheng.
\newblock {REFLECT: Intervention-Supported Error Attribution for Silent
  Failures in LLM Agent Traces}.
\newblock \emph{arXiv preprint arXiv:2606.09071}, 2026.

\bibitem[Liu et~al.(2026{\natexlab{a}})Liu, Xi, Zhang, Zeng, Yue, Wang, Kang,
  Wu, and Wang]{liu2026whowhenpro}
Jiale Liu, Huajun Xi, Shaokun Zhang, Yifan Zeng, Tianwei Yue, Chi Wang, Jian
  Kang, Qingyun Wu, and Huazheng Wang.
\newblock {Who\&When Pro: Can LLMs Really Attribute Failures in AI Agents?}
\newblock \emph{arXiv preprint arXiv:2607.09996}, 2026{\natexlab{a}}.

\bibitem[Liu et~al.(2026{\natexlab{b}})Liu, Li, Du, Luo, Zeng, Tewolde, Lee,
  Wang, Kingsford, and Conitzer]{liu2026memorycurse}
Jiayuan Liu, Tianqin Li, Shiyi Du, Xin Luo, Haoxuan Zeng, Emanuel Tewolde,
  Tai~Sing Lee, Tonghan Wang, Carl Kingsford, and Vincent Conitzer.
\newblock {The Memory Curse: How Expanded Recall Erodes Cooperative Intent in
  LLM Agents}.
\newblock \emph{arXiv preprint arXiv:2605.08060}, 2026{\natexlab{b}}.

\bibitem[Ma et~al.(2025)Ma, Zhang, Yang, Kang, Lin, Rajmohan, and
  Zhang]{ma2025dover}
Ming Ma, Jue Zhang, Fangkai Yang, Yu~Kang, Qingwei Lin, Saravan Rajmohan, and
  Dongmei Zhang.
\newblock {DoVer: Intervention-Driven Auto Debugging for LLM Multi-Agent
  Systems}.
\newblock \emph{arXiv preprint arXiv:2512.06749}, 2025.

\bibitem[Min et~al.(2026{\natexlab{a}})Min, Wu, Darbari, Chen, and
  Hong]{min2026trace}
Guanghui Min, Liang Wu, Mayank Darbari, Chen Chen, and Liangjie Hong.
\newblock {Toward Reliable Context Compression for Long-Horizon Agents: An
  Empirical Study of Execution Instability}.
\newblock \emph{arXiv preprint arXiv:2608.06503}, 2026{\natexlab{a}}.

\bibitem[Min et~al.(2026{\natexlab{b}})Min, Wu, Darbari, Chen, and
  Hong]{trace2026repo}
Guanghui Min, Liang Wu, Mayank Darbari, Chen Chen, and Liangjie Hong.
\newblock {TRACE: paired-continuation corpus and verifier}.
\newblock \url{https://github.com/nokia-applied-research/Trace},
  2026{\natexlab{b}}.
\newblock Revision 9be7d669.

\bibitem[O'Callahan et~al.(2017)O'Callahan, Jones, Froyd, Huey, Noll, and
  Partush]{ocallahan2017rr}
Robert O'Callahan, Chris Jones, Nathan Froyd, Kyle Huey, Albert Noll, and
  Nimrod Partush.
\newblock {Engineering Record and Replay for Deployability}.
\newblock In \emph{2017 USENIX Annual Technical Conference (USENIX ATC 17)},
  pages 377--389, 2017.
\newblock Extended technical report: arXiv:1705.05937.

\bibitem[Shah(2026)]{shah2026causalreplay}
Jaineet Shah.
\newblock {Causal Agent Replay: Counterfactual Attribution for LLM-Agent
  Failures}.
\newblock \emph{arXiv preprint arXiv:2606.08275}, 2026.

\bibitem[Sinha et~al.(2026)Sinha, Arun, Goel, Staab, and
  Geiping]{sinha2025illusion}
Akshit Sinha, Arvindh Arun, Shashwat Goel, Steffen Staab, and Jonas Geiping.
\newblock {The Illusion of Diminishing Returns: Measuring Long Horizon
  Execution in LLMs}.
\newblock In \emph{International Conference on Learning Representations
  (ICLR)}, 2026.
\newblock arXiv:2509.09677.

\bibitem[Song et~al.(2025)Song, Jayarajan, Ding, Su, Zhu, Liu, and
  Pekhimenko]{song2025aegis}
Kevin Song, Anand Jayarajan, Yaoyao Ding, Qidong Su, Zhanda Zhu, Sihang Liu,
  and Gennady Pekhimenko.
\newblock {Aegis: Taxonomy and Optimizations for Overcoming Agent-Environment
  Failures in LLM Agents}.
\newblock \emph{arXiv preprint arXiv:2508.19504}, 2025.

\bibitem[Trivedi et~al.(2024)Trivedi, Khot, Hartmann, Manku, Dong, Li, Gupta,
  Sabharwal, and Balasubramanian]{trivedi2024appworld}
Harsh Trivedi, Tushar Khot, Mareike Hartmann, Ruskin Manku, Vinty Dong, Edward
  Li, Shashank Gupta, Ashish Sabharwal, and Niranjan Balasubramanian.
\newblock {AppWorld: A Controllable World of Apps and People for Benchmarking
  Interactive Coding Agents}.
\newblock \emph{Proceedings of ACL 2024; arXiv:2407.18901}, 2024.

\bibitem[Vig et~al.(2020)Vig, Gehrmann, Belinkov, Qian, Nevo, Sakenis, Huang,
  Singer, and Shieber]{vig2020causal}
Jesse Vig, Sebastian Gehrmann, Yonatan Belinkov, Sharon Qian, Daniel Nevo,
  Simas Sakenis, Jason Huang, Yaron Singer, and Stuart Shieber.
\newblock {Causal Mediation Analysis for Interpreting Neural NLP: The Case of
  Gender Bias}.
\newblock \emph{Advances in Neural Information Processing Systems 33 (NeurIPS
  2020); arXiv:2004.12265}, 2020.

\bibitem[Wang et~al.(2026)Wang, Bai, Sun, Wang, Zhang, Hu, Schroder, Mutlu,
  Song, and Nowak]{wang2026horizon}
Xinyu~Jessica Wang, Haoyue Bai, Yiyou Sun, Haorui Wang, Shuibai Zhang, Wenjie
  Hu, Mya Schroder, Bilge Mutlu, Dawn Song, and Robert~D. Nowak.
\newblock {The Long-Horizon Task Mirage? Diagnosing Where and Why Agentic
  Systems Break}.
\newblock \emph{arXiv preprint arXiv:2604.11978}, 2026.

\bibitem[Xia et~al.(2026)Xia, Gao, Quan, Liu, and Fang]{xia2026agentlocate}
Yufei Xia, Anjun Gao, Yueyang Quan, Zhuqing Liu, and Minghong Fang.
\newblock {Who Broke the System? Failure Localization in LLM-Based Multi-Agent
  Systems}.
\newblock \emph{arXiv preprint arXiv:2607.07989}, 2026.

\bibitem[Xiao et~al.(2026)Xiao, Wang, Chen, Chao, Cui, Qian, Meng, Mei, Jiang,
  Ouyang, and Yi]{xiao2026errorsmemories}
Shuyao Xiao, Shengling Wang, Xuan Chen, Ke~Chao, Ming Cui, Feifei Qian, Fanlin
  Meng, Chaoyang Mei, Chaoyong Jiang, Qi~Ouyang, and Junxi Yi.
\newblock {When Errors Become Memories: Causal Pathway Tracing in Multi-Turn
  Memory-Augmented LLMs}.
\newblock \emph{arXiv preprint arXiv:2608.30198}, 2026.

\bibitem[Yang et~al.(2026)Yang, Song, Li, Ganguly, Ma, Wang, Dou, Zhou,
  Chaudhary, and Han]{yang2026agentce}
Wang Yang, Chaoda Song, Xinpeng Li, Debargha Ganguly, Chuang Ma, Shouren Wang,
  Zhihao Dou, Yuli Zhou, Vipin Chaudhary, and Xiaotian Han.
\newblock {AgentCE-Bench: Agent Configurable Evaluation with Scalable Horizons
  and Controllable Difficulty under Lightweight Environments}.
\newblock \emph{arXiv preprint arXiv:2604.06111}, 2026.

\bibitem[Yao et~al.(2026)Yao, Liu, Chen, Chen, Chen, He, Ni, Wang, and
  Ren]{yao2026actbench}
Hongwei Yao, Yiming Liu, Meihui Chen, Jieling Chen, Zikun Chen, Yiling He,
  Wangze Ni, Cong Wang, and Kui Ren.
\newblock {ActBench: Self-Evolving Benchmark of Behavioral Safety in Cowork
  Agents}.
\newblock \emph{arXiv preprint arXiv:2608.09476}, 2026.

\bibitem[Yuan et~al.(2025)Yuan, Li, Ding, Xie, Li, Zhao, Wan, Shi, Hu, and
  Liu]{yuan2025numerical}
Jiayi Yuan, Hao Li, Xinheng Ding, Wenya Xie, Yu-Jhe Li, Wentian Zhao, Kun Wan,
  Jing Shi, Xia Hu, and Zirui Liu.
\newblock {Understanding and Mitigating Numerical Sources of Nondeterminism in
  LLM Inference}.
\newblock \emph{arXiv preprint arXiv:2506.09501}, 2025.

\bibitem[Zhang et~al.(2025{\natexlab{a}})Zhang, Wang, Chen, Zhou, Wang, and
  Yan]{zhang2025agentracer}
Guibin Zhang, Junhao Wang, Junjie Chen, Wangchunshu Zhou, Kun Wang, and
  Shuicheng Yan.
\newblock {AgenTracer: Who Is Inducing Failure in the LLM Agentic Systems?}
\newblock \emph{arXiv preprint arXiv:2509.03312}, 2025{\natexlab{a}}.

\bibitem[Zhang et~al.(2026)Zhang, Wang, Sun, Pei, Sun, Xie, and
  Li]{zhang2026duotrace}
Jiayi Zhang, Zexin Wang, Degang Sun, Changhua Pei, Fei Sun, Gaogang Xie, and
  Jingjing Li.
\newblock {Detect Before You Attribute: Cascade Failure Attribution for
  Multi-Agent Systems}.
\newblock \emph{arXiv preprint arXiv:2608.29646}, 2026.

\bibitem[Zhang et~al.(2025{\natexlab{b}})Zhang, Yin, Zhang, Liu, Han, Zhang,
  Li, Wang, Wang, Chen, and Wu]{zhang2025whowhen}
Shaokun Zhang, Ming Yin, Jieyu Zhang, Jiale Liu, Zhiguang Han, Jingyang Zhang,
  Beibin Li, Chi Wang, Huazheng Wang, Yiran Chen, and Qingyun Wu.
\newblock {Which Agent Causes Task Failures and When? On Automated Failure
  Attribution of LLM Multi-Agent Systems}.
\newblock \emph{arXiv preprint arXiv:2505.00212}, 2025{\natexlab{b}}.

\bibitem[Zhang et~al.(2023)Zhang, Guo, Stepputtis, Sycara, and
  Campbell]{zhang2023understanding}
Xijia Zhang, Yue Guo, Simon Stepputtis, Katia Sycara, and Joseph Campbell.
\newblock {Understanding Your Agent: Leveraging Large Language Models for
  Behavior Explanation}.
\newblock \emph{arXiv preprint arXiv:2311.18062}, 2023.

\bibitem[Zhu et~al.(2025{\natexlab{a}})Zhu, Hong, Wu, Chawla, Tang, Yin, Wolfe,
  Babinsky, and Liu]{zhu2025raffles}
Chenyang Zhu, Spencer Hong, Jingyu Wu, Kushal Chawla, Charlotte Tang, Youbing
  Yin, Nathan Wolfe, Erin Babinsky, and Daben Liu.
\newblock {RAFFLES: Reasoning-based Attribution of Faults for LLM Systems}.
\newblock \emph{arXiv preprint arXiv:2509.06822}, 2025{\natexlab{a}}.

\bibitem[Zhu et~al.(2025{\natexlab{b}})Zhu, Liu, Li, Tian, Yang, Zhang, Han,
  Xie, Cui, Zhang, Ma, Yu, Ramesh, Wu, Liu, Lu, Zou, and
  You]{zhu2025agentdebug}
Kunlun Zhu, Zijia Liu, Bingxuan Li, Muxin Tian, Yingxuan Yang, Jiaxun Zhang,
  Pengrui Han, Qipeng Xie, Fuyang Cui, Weijia Zhang, Xiaoteng Ma, Xiaodong Yu,
  Gowtham Ramesh, Jialian Wu, Zicheng Liu, Pan Lu, James Zou, and Jiaxuan You.
\newblock {Where LLM Agents Fail and How They can Learn From Failures}.
\newblock \emph{arXiv preprint arXiv:2509.25370}, 2025{\natexlab{b}}.

\end{thebibliography}

\appendix

\IfFileExists{appendix/appA-carrier.tex}{\section{Measurement 3.1: the carrier taxonomy}\label{app:carrier}

\subsection{Corpus and carrier labels}

The corpus is a released set of paired benign and adversarial agent
trajectories \citep{yao2026actbench}: 300 task pairs and 24{,}000
trajectories over 20 released model--harness configurations (15 models, 6 harnesses; not a full cross), all parsed
without error.

Carrier labels are deterministic: the rule reads the benchmark's primary-carrier metadata (a mock-fixture field maps to \texttt{service\_response}, a workspace file to the local classes), gated on the SHA-256-changed sources of the benign and adversarial bundles (215 and 36 pairs); where that metadata was absent, a deterministic path-prefix rule over the changed sources labeled 15 further pairs. The rule fired on 266/300 task pairs (88.67\%), above its pre-specified floor of 85\%; the 34 unlabeled pairs, which lacked usable metadata and had both fixture and workspace sources changed, and their 2{,}040 attack trajectories were excluded rather than assigned non-deterministically.
Fine-class counts before collapse were \texttt{service\_response} 225, \texttt{memory\_state} 15,
\texttt{tool\_metadata} 14 and \texttt{workspace\_file} 12. The protocol
required 30 task pairs per analyzed class and pre-specified that classes short of it would be collapsed into a coarser taxonomy;
the specific service-versus-local mapping was chosen at collapse time, after the fine-class counts were known. All three local classes fell short and were merged into
\texttt{local\_file\_state}---task-reachable workspace documents,
persistent memory state, and tool or skill metadata---leaving
\texttt{service\_response} (225 pairs) and \texttt{local\_file\_state} (41).
The merged class pools carriers with different recovery rates, so the collapsed contrast supports no inference about memory, tool-metadata or workspace carriers separately.

\subsection{Audit protocol}

Fifty task pairs were relabeled by the same LLM agent that ran the gate, not blind to the rule (each packet displayed the rule's carrier label and the benchmark's primary-carrier field), with
raw-bundle access: it read each case's benign/adversarial source diff and
task metadata and recorded a coarse class, a fine class and a note.
Agreement with the deterministic rule was 50/50, against a pre-specified
floor of 45/50; the sample split 44 service to 6 local, and 44/2/2/2 fine.

\subsection{Two pre-results additions to the protocol}

Both were logged after the 50-case tallies had been seen and before any
concentration, odds-ratio, exposure or replication number existed; the skew
of the audit sample was the sole trigger. Paraphrased: first, if fewer than three
classes clear the cell-size floor after the collapse, the concentration and
ranking layers are not evaluable---a top-two share over two exhaustive
classes is 1.0 by construction and a rank correlation over two items can
only be $\pm 1$---leaving the exposure-conditional layer, which conditions
on exposures rather than on the carrier marginal. Second, first-exposure
recovery must be reported per analyzed class, each class clearing 70\%: a
class below that floor estimates its conversion rate on a differently
biased subsample, so the cross-class comparison would be confounded by
recoverability rather than by behavior.

\subsection{Estimates}

Table~\ref{tab:carrier-main} reports the analyzed estimates, all computed
on the 15{,}960 labeled attack trajectories from the 266 labeled pairs
under the benchmark's own unsafe-action label.

\begin{table}[h]
\centering
\small
\caption{Measurement 3.1, analyzed carrier estimates. Intervals are 95\% task-clustered bootstrap intervals at seed 20260901 (within-scenario dependence between tasks is not represented), with 5{,}000 replicates for shares and concentration indices and 2{,}000 for the adjusted model. Attack method is nearly collinear with carrier (only context injection contains both classes: 195 service, 12 local pairs), so the adjusted odds ratio is estimated under limited common support. The adjusted odds ratio is a logistic carrier contrast with fixed effects for 15 risk-behavior classes and the benchmark's attack method, fit on all 15{,}960 labeled rows, with 0 failed bootstrap fits.}
\label{tab:carrier-main}
\begin{tabular}{@{}lrr@{}}
\toprule
Quantity & \texttt{service\_response} & \texttt{local\_file\_state} \\
\midrule
Task pairs (of 266 labeled) & 225 (84.6\%) & 41 (15.4\%) \\
Labeled attack trajectories & 13{,}500 & 2{,}460 \\
Attack-positive trajectories (of 9{,}199) & 7{,}810 & 1{,}389 \\
Share of unsafe actions & 0.8490 [0.8008, 0.8940] & 0.1510 [0.1060, 0.1992] \\
Adjusted odds ratio & 1.1484 [0.6275, 2.0160] & 1 (reference) \\
First-exposure recovery & 13{,}048/13{,}500 = 96.65\% & 1{,}705/2{,}460 = 69.31\% \\
Exposure-conditional conversion & 7{,}629/13{,}048 = 58.47\% & 1{,}122/1{,}705 = 65.81\% \\
\bottomrule
\end{tabular}

\smallskip
{\footnotesize First exposure is the earliest transcript message containing
an injected candidate or a fragment of it; for service fixtures the candidate set holds the entire attacked field value, so a fragment match can fall on unchanged benign text, whereas local-file candidates are the changed spans only.
Pooled: 14{,}753/15{,}960 first exposures located (92.44\%), payload
candidates recoverable for 15{,}960/15{,}960; HHI 0.7436 [0.6809, 0.8104],
Gini 0.3490 [0.3008, 0.3940]; top-one share half-width 0.0466; top-two
share 1.0000, half-width 0.0000---mechanical at two exhaustive classes.}
\end{table}

\subsection{Replication and sensitivity}

Table~\ref{tab:carrier-sens} records the folds for completeness, not as
evidence. In the clean-twin control, the benign halves of the same pairs already
differ by carrier class, on an outcome that no payload touches.

\begin{table}[h]
\centering
\small
\caption{Replication folds and sensitivity analyses; same seed, 2{,}000 replicates for the proxy model. Leave-one-out replication ran on behavior- and attack-method-adjusted carrier log-odds rankings along both released axes; with two surviving classes $\tau$-b can take only $\pm 1$.}
\label{tab:carrier-sens}
\begin{tabular}{@{}lp{0.6\textwidth}@{}}
\toprule
Analysis & Result \\
\midrule
Leave-one-model-out, harness fixed & 15/15 computable folds; $\tau$-b median 1.000, minimum $-1.000$ (14 folds at $+1$, one at $-1$) \\
Leave-one-harness-out, model fixed & 6/6 computable folds; $\tau$-b median 1.000, minimum 1.000 \\
Per-configuration share matrix & 20 configurations $\times$ 2 classes, 40 cells \\
Log-evidence-only proxy & 7{,}150 events; service share 0.8236; adjusted OR 0.8973 [0.4715, 1.7043] \\
Clean-twin negative control & benign-task failure 13.49\% (607/4{,}500) for service twins vs.\ 8.78\% (72/820) for local twins; 5{,}320 rows, 266 pairs \\
Drop configurations below 200 usable & all 20 had 798 usable labeled attack trajectories; 0 dropped, estimates unchanged \\
\bottomrule
\end{tabular}
\end{table}

\subsection{Disposition under the frozen rule}

The measurement was declared non-evaluable under the pre-outcome amended criterion. Two classes survived the collapse, so the
concentration and ranking layers were not evaluable; and
\texttt{local\_file\_state} recovery of 69.31\% missed the per-class floor
of 70\% by 17 trajectories, leaving one evaluable class where the protocol
required two, so the exposure-conditional layer failed too. The failure is substantive rather than an artifact of the threshold. Of the trajectories whose first exposure was not located, 181/452 (40.0\%) for
\texttt{service\_response} and 267/755 (35.4\%) for \texttt{local\_file\_state}
were nonetheless attack-positive under the benchmark label, despite having no located first exposure. The
located rate is therefore the generous reading of recovery, and the 27.3-point
differential is the confound the per-class floor was written to catch.
}{}
\IfFileExists{appendix/appB-replay.tex}{\section{Measurement 3.2: replay fidelity}\label{app:replay}

\subsection{Census and controls}

The environment is AgentCE-Bench \citep{yang2026agentce} at repository revision \texttt{22ae16b7}, under our own snapshot layer
(the benchmark ships no fork or restore API); a certificate should state its population of attempted restorations and count aborted or unreachable ones; here a snapshot is certified by restoring five
clones and comparing SHA-256 digests. Census
(seed 424242, tool-failure channel off):
144 verdicts over 108 distinct states (see below) $=$ 6 domains $\times$ 3 grids ($5\times5$, $5\times7$, $5\times10$)
$\times$ 2 horizons (lowest and highest hidden-slot count, 1--25)
$\times$ 2 branch budgets ($0$, $8$) $\times$ 2 anchors ($1/3$, $2/3$);
144/144 pass, failure fraction $0$; the 144 verdicts cover 108 distinct states (the 72 one-slot rows are measured at both
anchors, which coincide). A further 17 model-generated states passed the same gate.
Two controls bound what the certificate can see (an in-snapshot sensitivity check cannot, by
construction, detect omitted state): three single-field mutations
(query-counter increment, grid-cell change, dropped last message) injected into
15 states were detected 45/45 (seed 20260901),
and 20 states restored by five concurrent separate processes passed 20/20
(seed 20260902).

\subsection{The uncaptured stream}

With a positive tool-failure rate, the environment's tool-dispatch layer draws
every failure from the process-global \texttt{random} stream. The benchmark's
entry point seeds that stream at start-up and re-seeds it while constructing
the task cache; a harness that constructs tasks directly, as ours does, does
neither.
Neither restores the stream's \emph{position},
and no snapshot field records it, so digest-identical clones resume at different
points of one stream. Our
repair is harness-side: capture the stream state at snapshot construction,
restore it before each continuation; no upstream file or snapshot schema was
modified.

\subsection{Divergence measurements}

Design: 24 states $=$ 6 domains $\times$ hidden-slot count $\in \{5, 21\}$ $\times$
anchors $1/3$, $2/3$, on the $5\times7$ grid at branch budget 8;
deterministic prefix at rate $0$, treatment rate from the anchor onward, fixed
ground-truth continuation policy, at most 10 attempts per slot.
Observables: per-call success/failure sequence, final serialized-snapshot hash (it includes the transcript, so it is not independent of the failure pattern), final grader
result. All 120 clones in each naive five-clone condition restored to the
candidate digest;
51 tool calls failed at rate $0.05$, 229 at $0.20$.
Table~\ref{tab:appb-rng} gives every measured row.
\clearpage

\begin{table}[p]
\centering
\footnotesize
\caption{Replay divergence in the unseeded experiment, all measured conditions (B1: five-clone experiment; B2: live-vs-restored and live-vs-live experiment; the seeded replication is in Table~\ref{tab:appb-b4}). ``Naive'' restores the
serialized state only; ``captured'' additionally restores the random-stream
position; live-vs-live rows restore nothing and are the unseeded re-execution baseline of the run named in the Machinery column. 10{,}000 state resamples per interval, seed 20260901; every fraction
and interval endpoint is an integer multiple of $1/24$ and is shown to two
decimals. Intervals are the released state-resampling bootstrap intervals over the 24 factorial cells and are descriptive: the cells are not a random sample of the census, so no population coverage is claimed; at a zero count the bootstrap is degenerate (a binomial Clopper--Pearson 95\% upper limit would be 0.14~\citep{clopper1934}, under a sampling model the design does not supply).}
\label{tab:appb-rng}
\begin{tabular}{llllrl}
\toprule
Exp. & Rate & Machinery & Comparison / metric & Div./24 & Fraction [95\% CI] \\
\midrule
B1 & 0.05 & naive & five-clone / observable & 17 & 0.71 [0.54, 0.88] \\
B1 & 0.20 & naive & five-clone / observable & 22 & 0.92 [0.79, 1.00] \\
B2 & 0.05 & naive & live vs.\ restored / observable & 11 & 0.46 [0.25, 0.67] \\
B2 & 0.05 & naive & live vs.\ restored / failure pattern & 11 & 0.46 [0.25, 0.67] \\
B2 & 0.05 & naive & live vs.\ restored / final hash & 11 & 0.46 [0.25, 0.67] \\
B2 & 0.05 & naive & live vs.\ restored / final grader & 0 & 0.00 [0.00, 0.00] \\
B2 & 0.05 & naive & live vs.\ live / observable & 13 & 0.54 [0.33, 0.75] \\
B2 & 0.05 & naive & live vs.\ live / failure pattern & 13 & 0.54 [0.33, 0.75] \\
B2 & 0.05 & naive & live vs.\ live / final hash & 13 & 0.54 [0.33, 0.75] \\
B2 & 0.05 & naive & live vs.\ live / final grader & 0 & 0.00 [0.00, 0.00] \\
B2 & 0.20 & naive & live vs.\ restored / observable & 19 & 0.79 [0.63, 0.96] \\
B2 & 0.20 & naive & live vs.\ restored / failure pattern & 19 & 0.79 [0.63, 0.96] \\
B2 & 0.20 & naive & live vs.\ restored / final hash & 19 & 0.79 [0.63, 0.96] \\
B2 & 0.20 & naive & live vs.\ restored / final grader & 0 & 0.00 [0.00, 0.00] \\
B2 & 0.20 & naive & live vs.\ live / observable & 18 & 0.75 [0.58, 0.92] \\
B2 & 0.20 & naive & live vs.\ live / failure pattern & 18 & 0.75 [0.58, 0.92] \\
B2 & 0.20 & naive & live vs.\ live / final hash & 18 & 0.75 [0.58, 0.92] \\
B2 & 0.20 & naive & live vs.\ live / final grader & 0 & 0.00 [0.00, 0.00] \\
\midrule
B1 & 0.05 & captured & five-clone / observable & 0 & 0.00 [0.00, 0.00] \\
B1 & 0.20 & captured & five-clone / observable & 0 & 0.00 [0.00, 0.00] \\
B2 & 0.05 & captured & live vs.\ restored / observable & 0 & 0.00 [0.00, 0.00] \\
B2 & 0.05 & captured & live vs.\ restored / failure pattern & 0 & 0.00 [0.00, 0.00] \\
B2 & 0.05 & captured & live vs.\ restored / final hash & 0 & 0.00 [0.00, 0.00] \\
B2 & 0.05 & captured & live vs.\ restored / final grader & 0 & 0.00 [0.00, 0.00] \\
B2 & 0.05 & captured & live vs.\ live / observable & 10 & 0.42 [0.21, 0.63] \\
B2 & 0.05 & captured & live vs.\ live / failure pattern & 10 & 0.42 [0.21, 0.63] \\
B2 & 0.05 & captured & live vs.\ live / final hash & 10 & 0.42 [0.21, 0.63] \\
B2 & 0.05 & captured & live vs.\ live / final grader & 0 & 0.00 [0.00, 0.00] \\
B2 & 0.20 & captured & live vs.\ restored / observable & 0 & 0.00 [0.00, 0.00] \\
B2 & 0.20 & captured & live vs.\ restored / failure pattern & 0 & 0.00 [0.00, 0.00] \\
B2 & 0.20 & captured & live vs.\ restored / final hash & 0 & 0.00 [0.00, 0.00] \\
B2 & 0.20 & captured & live vs.\ restored / final grader & 0 & 0.00 [0.00, 0.00] \\
B2 & 0.20 & captured & live vs.\ live / observable & 18 & 0.75 [0.58, 0.92] \\
B2 & 0.20 & captured & live vs.\ live / failure pattern & 18 & 0.75 [0.58, 0.92] \\
B2 & 0.20 & captured & live vs.\ live / final hash & 18 & 0.75 [0.58, 0.92] \\
B2 & 0.20 & captured & live vs.\ live / final grader & 0 & 0.00 [0.00, 0.00] \\
\bottomrule
\end{tabular}
\end{table}

The unseeded five-clone figures (17/24, 22/24) are in the direction and magnitude given by the analytic expectation (0.73 at rate 0.05, 0.98 at 0.20; 17.4 and 23.6 of 24 expected against 17 and 22 observed) for five independent draw sequences computed from that experiment's own mean clone lengths, a post hoc plug-in check from the run's own observed lengths, not a prediction;
the result is that the hash certificate was blind to them.
The live-vs-live arm is an unseeded re-execution baseline, not a treatment; it moved between
runs at rate $0.05$ (13/24, then 10/24) and sat at 18/24 at $0.20$ under both machineries.
The seeded replication below removes that baseline's noise: two fresh single-threaded seeded
re-executions agree in 24/24 states (the upstream's shipped runner executes tasks in a thread
pool over one shared stream, a further uncaptured channel).
The unseeded live-vs-restored figures in Table~\ref{tab:appb-rng} (11/24, 19/24) are from this
run; the replication's unseeded arm is a separate run (14/24, 21/24).

\subsection{Grader agreement}

No five-clone condition held a state with more than one distinct final grader
result: 0 of 24 at each rate, under both machineries.
Under this retry-until-success oracle policy, realization-level divergence never
flipped the outcome; an outcome-only metric would have reported perfect
fidelity.

\subsection{A second environment}

AppWorld 0.1.3.post1 \citep{trivedi2024appworld}.
Reset probe: 10 tasks $\times$ 5 fresh initializations compared on an
independent full-database digest, the benchmark's row-hash digest, a frozen
clock, and untouched-state evaluator output---0/10 mismatched, 50/50
identical.
Replay probe: 5 tasks, scripted prefix with one state-changing write, full
per-application SQLite snapshot, 5 restored copies, two-read suffix---0/5
mismatched, 25/25 identical.
The checkpoint/replay probe required first resetting cached named in-memory database engines;
otherwise a file-copy restore was silently ignored, with no error raised, and was caught only
by the probe's restore-identity digest.
AppWorld ships \texttt{save\_state}/\texttt{load\_state}, which we did not
evaluate: the lesson is scoped to harness authors who build their own
file-level snapshots.

\clearpage
\subsection{Seeded replication}

A second run under a pre-registration frozen before execution, and amended as disclosed below, re-executed everything seed-once: trajectories run at the treatment rate from the start, compared with the live continuation, on the same 24 states.
Two fresh runs with the same seed agree exactly, and so does a state rebuilt by replaying the recorded prefix; a byte-identical snapshot restored into a freshly seeded process does not, under either seed regime, and five such clones agree with one another in 24/24 states while differing from the live run in 9 and 20 of 24 with per-state seeds (18 and 24 of 24 under the runner's fixed seed, which yields only four distinct live realizations across the 24 states) (Table~\ref{tab:appb-b4}, Figure~\ref{fig:appb-b4}).
A clone-vs-clone gate whose clones share a seed position therefore passes and is still wrong; a comparison against a separately executed reference continuation, live or prefix-rebuilt, detects it. The independent-sequence references quoted in the main text (0.45, 0.84) were recomputed with the pre-registered formula from the per-state run's live continuation lengths (mean 7.04 and 8.13 calls); the released expectation file, computed from the fixed-seed run's lengths, gives 0.47 and 0.83. They are plug-in references, not the exact null for this design: the live and restored continuations read shifted portions of one per-state seeded stream, and the realized live lengths depend on the retry rule.

Two arms were added by dated amendment after the fixed-seed results had been seen: the five-clone seeded arm, to test the clone-agreement consequence that result suggested, and the per-state-seed repeat, because one process seed makes the six domains of a cell share a single live realization (four distinct realizations across the 24 states). No acceptance criterion changed; the per-state arms were designated primary after both result sets were known; the pre-registration file's own dated correction records that its amendment timestamps had been misstated by five to eight minutes.

\begin{table}[h]
\centering
\footnotesize
\caption{Seeded replication: states (of 24) whose continuation diverged from the live continuation, by arm and tool-failure rate. Counts only; the 24 states are factorial cells. The last two rows report clone-vs-clone disagreement (both seed regimes) and final-grader divergence.}
\label{tab:appb-b4}
\begin{tabular}{@{}lll@{}}
\toprule
Arm & 0.05 & 0.20 \\
\midrule
Two fresh runs, same seed (re-execution baseline) & 0/24 & 0/24 \\
State rebuilt by replaying the recorded prefix (seeded) & 0/24 & 0/24 \\
Byte-identical snapshot restored, seeded process, runner's fixed seed (four distinct realizations) & 18/24 & 24/24 \\
Byte-identical snapshot restored, seeded process, one seed per state & 9/24 & 20/24 \\
Byte-identical snapshot restored, unseeded process (separate run) & 14/24 & 21/24 \\
Snapshot with captured random-stream state restored & 0/24 & 0/24 \\
\midrule
Five restored clones, seeded process: disagreement among the clones & 0/24 & 0/24 \\
Final-grader divergence, any arm & 0/24 & 0/24 \\
\bottomrule
\end{tabular}
\end{table}

\begin{figure}[h]
\centering
\includegraphics[width=0.9\linewidth]{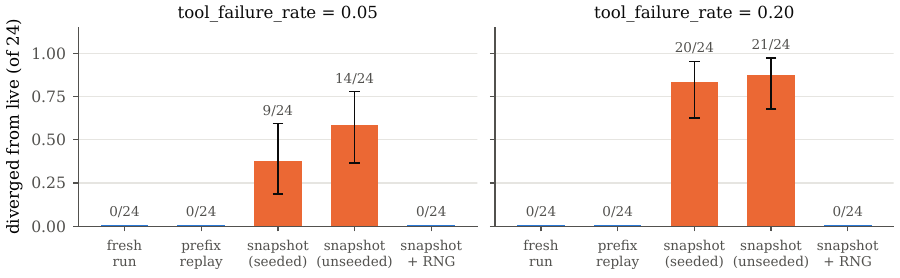}
\caption{Seeded replication, one seed per state: divergence from the live continuation (fraction of 24 states) for the arms of Table~\ref{tab:appb-b4} at tool-failure rate 0.05 (left) and 0.20 (right); error bars are exact binomial 95\% intervals drawn for orientation only, since the states are factorial cells; the unseeded bar is the separate unseeded run of Table~\ref{tab:appb-b4}.}
\label{fig:appb-b4}
\end{figure}

\subsection{A provider-side observation}

On $\tau^2$-bench \citep{barres2025tau2}, with a third-party-served model behind
a proxy at temperature $1.0$ and a best-effort seed,
three request payloads were each re-sent byte-identically with the same seed (four comparable call pairs across the three), and none reproduced its completion (0 of 4), while the serving path we controlled reproduced 34/34 calls and 2/2 episodes bitwise.
Temperature-zero decoding is not reproducible across runs in general~\citep{atil2024nondeterminism}, and greedy decoding diverges across batch sizes and hardware under 16-bit precision~\citep{yuan2025numerical}, so a hosted pathwise certificate must be measured on the deployed configuration. 52 of 56 regeneration calls had left the original key path, so live
re-execution yields about one comparable call per episode.
No response carried a \texttt{system\_fingerprint}, so silent downstream discard
of the seed is not excluded; environment-hash identity there carried
approximately zero discriminating information and is not offered as replay
evidence.
The sample is small, and non-reproduction is localized to the hosted path; we
cannot localize it further within that path.
}{}
\IfFileExists{appendix/appC-trace.tex}{\section{TRACE reproduction and a measurement not reported here}\label{app:trace}

\subsection{What was reproduced}

The TRACE release \citep{min2026trace,trace2026repo} is a frozen AppWorld cohort
\citep{trivedi2024appworld}: 590 compaction boundaries, 4{,}641 paired PRE/POST
rollouts (4{,}640 scored), 20{,}133 steps, 147 tasks (140 with boundaries). Its
shipped paired burden (TRACE's per-boundary paired union delta $d_U$ between its two arms) rises from $+0.131$ $[+0.101, +0.160]$ at one action after
a boundary to $+0.509$ $[+0.432, +0.588]$ at five. We re-ran the
released scan read-only at revision \texttt{9be7d669}, modifying no data: unit
tests 24/24 and 35/35, reproduction suite 4/6, a fresh scan over the same 590
boundaries and 20{,}133 steps, four of six shipped tables byte-identical.

\subsection{The byte-identity failure and the amended criterion}

Our frozen criterion required every shipped table to match by SHA-256. It failed on the two tables carrying bootstrap intervals, and on no others. An amendment was logged after that failure was observed, on the recorded but mistaken ground that the bootstrap was unseeded and byte identity across platforms therefore over-strict; the bootstrap is seeded, and the discrepancy later disappeared on a supported interpreter, consistent with an interpreter-version explanation (below).
The amended criterion requires deterministic tables to be byte-identical and, on
the stochastic tables, identical point estimates with row-by-row interval overlap
and unchanged classification of whether an interval excludes zero
(Table~\ref{tab:bootstrap-tables}); differences below $10^{-12}$ counted as
negligible, three orders of magnitude above the largest observed difference. Downstream work used the fresh
scan as its basis.

\begin{table}[h]
\centering
\small
\caption{Shipped versus fresh bootstrap-bearing tables. All 330 intervals
overlap, and no interval changes its qualitative classification.}
\label{tab:bootstrap-tables}
\begin{tabular}{lrr}
\toprule
& \texttt{curves.csv} & \texttt{marginal.csv} \\
\midrule
Rows matched & 285/285 & 45/45 \\
Point-estimate values compared & 1{,}710 & 45 \\
Max $|$shipped $-$ fresh$|$, point estimates & $1.33\times10^{-15}$ & $1.53\times10^{-16}$ \\
Interval endpoints compared & 570 & 90 \\
Max $|$shipped $-$ fresh$|$, endpoints & $7.22\times10^{-16}$ & $2.78\times10^{-16}$ \\
Non-overlapping intervals & 0 & 0 \\
Qualitative-classification changes & 0 & 0 \\
\bottomrule
\end{tabular}
\end{table}

\subsection{Why the differences are not a result}

The released bootstrap is seeded---a fixed constant, 20260803, in the verifier
module, used by the resamplers---so these are not resampling differences. Our
re-run used Python 3.10.8 while the package declares Python $\ge$ 3.11, and
differences at $10^{-15}$ are consistent with floating-point summation order
across interpreter versions. We did not isolate the summation path ourselves. A subsequent re-run under
Python 3.12.13, an interpreter the package supports, reproduced all six shipped
tables byte-identically and passed the release's 6/6 reproduction checks, so the original criterion is met on a supported interpreter, consistent with the interpreter-version explanation.

\subsection{Prefix re-execution as a restored state: what the recipe hash covers}

TRACE does not snapshot: each rollout opens a fresh environment, re-executes the
recorded code prefix with no model in the loop, installs one context (PRE or
POST), and free-runs at most five primary actions. The stated reason is that the
environment's own \texttt{save\_state}/\texttt{load\_state} does not round-trip
API-server state. Restoration is checked instead by the recipe hash \texttt{replay\_identity}
$=$ \textsc{sha256}(task, split, boundary\_step, prefix\_codes), which both arms
of a boundary must carry identically.

Measurement 3.2 shows such a hash is not sufficient in general. The protocol's own rationale is that a nondeterministic prefix would surface as a divergence; an
input hash cannot detect that (identical hashes, different worlds). Two weaker grounds remain for treating the paired arms as comparable: no stochastic
tool-failure channel of the kind that drives Measurement 3.2 is exercised in this protocol, and we
found the surface deterministic under reset (ten tasks) and under mid-trajectory checkpoint/replay on five short
scripted prefixes, not TRACE's own (0/10 and 0/5 tasks with any mismatch, Appendix~\ref{app:replay}),
where a naive file-copy restore was nonetheless found to reuse cached in-process engines silently.
Whether the 590 reached states were identical was measured neither by the authors nor by us: TRACE
is the closest executable-surface precedent, and it reports no measured replay certificate, so its pairing is unverified.

\subsection{A prefix taxonomy not reported here}

A third measurement was planned: an observational taxonomy of boundary prefixes,
contrasting continuation burden across classes. It is not reported here, on our own
artifact review, for two reasons. First, the label did not measure its
construct---prefixes classified as containing a state-changing write
predominantly contained authentication endpoints as their only non-read call, so
the label separated boundaries by whether the agent had yet logged into any
application (a trajectory-phase proxy). Second, a first agreement check re-applied the labeling rule to the same
endpoint-method map and was discarded as circular; a later map-blind re-labeling by two labelers of different model families agreed with the map on all 119 endpoint functions, which validates the map but not the construct. No estimate, interval, or
cell count from the taxonomy appears anywhere in this paper.

}{}
\IfFileExists{appendix/appD-factorial.tex}{\section{History arms and paired contrasts (design only)}\label{app:factorial}

The design proposed in the second agenda item has never been run: every
quantity below is a design parameter, not a result. Two local open-weight checkpoint pairs supply the powered lane (288 states on the main surface) and a half-grid replication of it; two smaller surfaces (36--48 states) replicate for sign; two hosted models form a frontier lane.

\subsection{Arms}

The design has eight arms (Table~\ref{tab:arms}): two length controls, four selected content-by-role cells from the crossing of \emph{error content} (healed, natural own, matched foreign, synthetic) with \emph{source role} (self, prior-agent wrapper) at a frozen executable state, and two fixed-packet provenance arms outside the four content levels.

\begin{table}[h]
\centering
\small
\caption{History arms. Only assistant-side text differs; the frozen executable
state (environment and process state), the user turns and all tool outputs are identical across arms.
Arms 3--6 instantiate four of the eight crossed content-by-role cells (healed and
synthetic content appear only under the self wrapper); arms 7--8 carry the
provenance contrast on a fixed packet outside the four content levels; arm~5 is the
natural-content, prior-agent cell and is not a clean provenance contrast.}
\label{tab:arms}
\begin{tabular}{@{}rp{2.05cm}p{2.55cm}p{1.75cm}p{4.35cm}@{}}
\toprule
\# & Arm & Error content & Source role & Isolates \\
\midrule
1 & Short verified state & none &---& fresh-context baseline \\
2 & Clean-long & inert, token/turn-matched & self & length tax $L$ (with 1) \\
\midrule
3 & Healed own & errors corrected in place & self & error-content counterfactual \\
4 & Untouched own & natural errors & self & $\Delta_{\mathrm{err}}$ (with 3) \\
5 & Natural foreign & matched real foreign errors & prior-agent & provenance $\times$ natural content \\
6 & Synthetic error & injected matched errors & self & reproduction of the synthetic-injection result \citep{sinha2025illusion} \\
7 & Byte-identical self & fixed error packet & self & $\Delta_{\mathrm{prov}}$ (with 8) \\
8 & Byte-identical foreign & same bytes & prior-agent & provenance with content held fixed \citep{chen2026selfcorrection} \\
\bottomrule
\end{tabular}
\end{table}

\subsection{Matching rules}

\begin{itemize}\setlength{\itemsep}{1pt}
\item Long-history arms are matched on tokenizer-specific token count and on assistant/tool-turn count; only the short-state control differs in length.
\item Assistant-side text only: transformations touching user or tool content are rejected automatically. Arms 7 and 8 are hash-verified and differ in the attributing wrapper alone.
\item A state is admitted only if five restored clones give identical state hashes and grader outcomes
\emph{and} an unedited-history restored continuation matches a separately executed live or prefix-rebuilt reference under the experiment's policy, seed, and provider regime, with the pathwise rate or distributional margin fixed in advance. Clone agreement alone is insufficient, since Measurement~3.2 found grader agreement across clones whose trajectories had diverged from the live reference; the snapshot must also carry process state (Appendix~\ref{app:replay}).
\item A frozen judge builds and verifies the arms but never scores an outcome; it must first reach $\ge 0.90$ precision and $\ge 0.85$ recall on 400 synthetic corruptions plus 200 blinded, human-labeled real spans.
\item Primary success is exact and programmatic; on a pre-specified quarter of states, handoffs (a continuation started by one model from a checkpoint written by its partner model) are run in both directions between the two local open-weight checkpoint partners and between the two hosted frontier models.
\end{itemize}

\subsection{Three non-additive probes}

Three paired contrasts probe where early$\to$late decay comes from, a \emph{length tax}, an \emph{error-content effect} and a \emph{provenance effect}, each identified on its own content distribution (Table~\ref{tab:estimands}); they do not share a target distribution and are not asserted to sum to the total decay, so no residual term is identified. Arm 5 changes content and role together, and arms 7--8 use a fixed packet outside the four content levels, so the provenance contrast is identified on its own distribution.
Table~\ref{tab:dominance} states what each probe's sign would imply. If none of the three probes moves success, the design has not located the decay; it does not bound untested interventions.

\begin{table}[h]
\centering
\small
\caption{Estimands. All are risk differences in percentage points of exact
continuation success, estimated pairwise within forked states and
cluster-bootstrapped over tasks, except $S$, which is dimensionless and reported
unclipped. The reproduction tests compare $\Delta_{\mathrm{foreign}}$ and
$\Delta_{\mathrm{synth}}$ against $\Delta_{\mathrm{err}}$ as $\pm 5$-point
equivalence tests (both interval bounds inside the margin), not as failures to reject. $S$ is secondary, reported
only when its denominator is at least 15 points (the floor assumed in the power sketch below) and its numerator sign-stable; the claim
that contaminated history explains less than half of decay requires the 95\%
upper bound below 0.5.}
\label{tab:estimands}
\begin{tabular}{@{}lll@{}}
\toprule
Symbol & Contrast & Arms \\
\midrule
$L$ & $p_{\text{short}} - p_{\text{clean-long}}$ & 1 vs.\ 2 \\
$\Delta_{\mathrm{err}}(d)$ & $p_{\text{healed}}(d) - p_{\text{own}}(d)$, $d \in \{\text{early}, \text{late}\}$ & 3 vs.\ 4 \\
$\Delta_{\mathrm{foreign}}$, $\Delta_{\mathrm{synth}}$ & $p_{\text{healed}} - p_{\text{foreign}}$, $p_{\text{healed}} - p_{\text{synth}}$ & 3 vs.\ 5, 3 vs.\ 6 \\
$\Delta_{\mathrm{prov}}$ & $p_{\text{byte-self}} - p_{\text{byte-foreign}}$ & 7 vs.\ 8 \\
$S$ & $[\Delta_{\mathrm{err}}(\text{late}) - \Delta_{\mathrm{err}}(\text{early})] \,/\, [p_{\text{own}}(\text{early}) - p_{\text{own}}(\text{late})]$ & 3, 4 \\
\bottomrule
\end{tabular}
\end{table}

\begin{table}[h]
\centering
\small
\caption{What each probe's sign would imply, and the repair it points to. Because each sign implies a different repair, the intended result is the set of contrasts, not the sign of any single one.}
\label{tab:dominance}
\begin{tabular}{@{}lp{7.2cm}p{3.3cm}@{}}
\toprule
Probe & Signature & Repair implied \\
\midrule
Error content & $\Delta_{\mathrm{err}}$(late) large; foreign and synthetic arms reproduce it inside the $\pm 5$-point margin; $\Delta_{\mathrm{prov}}$ small & history editing \\
Provenance & $\Delta_{\mathrm{prov}}$ large with the bytes held fixed; the reproduction tests fail & role-wrapper handling \\
Length tax & $L$ large; contaminated arms barely move once token and turn counts are matched & neither; the effect was context length \\
None of the three & no contrast estimated precisely enough to locate the decay (the design is not powered for a 5-point equivalence margin) & not located by this design \\
\bottomrule
\end{tabular}
\end{table}

\subsection{Power sketch}

Primary contrasts are within-state paired differences. On the powered lane, a
configurable-horizon and deterministically graded tool environment
\citep{yang2026agentce}, 288 paired states clustered in 144 tasks run at three
repetitions per arm.
With $p \approx 0.4$--$0.6$ at the anchors and within-task ICC $\approx 0.2$,
the standard error of an arm difference under a zero-within-state-correlation planning approximation is $\approx \sqrt{2p(1-p)/n_{\text{eff}}}$ with $n_{\text{eff}} \approx 400$--550, i.e.\ 3.0--3.5 points, so the minimum detectable effect at 80\% power, $\alpha = .05$ two-sided, is $(1.96 + 0.84)\,\mathrm{SE} \approx 8$--10 points.
The 5-point equivalence margins therefore require either $n_{\text{eff}} \approx 1{,}500$ effective pairs or a pilot-estimated within-state covariance, which may shrink or widen the paired standard error; an earlier draft of this design quoted 4--5 points, which does not follow from the formula above; as sized, the design is powered for the sign of the primary contrasts, not for the margins.
The two replication lanes (the two smaller surfaces, 36--48 states each) and the frontier lane (the two hosted models) are powered only for large effects on the same arithmetic (minimum detectable effects of roughly 20 points or more, revised from the 8--12 and 8--10 points an earlier version of this design quoted under the correlation assumption set aside above) and serve as sign checks.
With a denominator of at least 15 points and $\Delta_{\mathrm{err}}$ standard errors near 3 points, the standard error of $S$ is roughly 0.3 and a 95\% interval roughly $\pm 0.55$: too wide to place $S$ against 0.5 except at the extremes, which is why $S$ is secondary.
Variance components are pilot-estimated.

\subsection{Cells already occupied}

Three cells are partly occupied. Sinha et al.\ run healed and controlled-error
histories, but on a synthetic running-sum task with the errors always in the
assistant role: content varies only under the assistant role, so no provenance effect or content-by-provenance interaction is identified, and no tokenizer-exact length match is reported \citep{sinha2025illusion}. Chen et al.\ run the byte-identical role swap,
but on single claims in short math and logic traces, scoring correction behavior
rather than end-task success \citep{chen2026selfcorrection}. Min et al.\ occupy
the compaction cell at matched decision points on an executable surface
\citep{min2026trace}. What is missing is a matched comparison of content and source role at a frozen
executable state, with a length control and scoring on task success. The
design has not been run.
}{}
\IfFileExists{appendix/appE-census.tex}{\section{Literature census}\label{app:census}

\paragraph{Question.} Of the papers in our reading list whose primary contribution is
failure taxonomy, attribution, localization or empirical diagnosis for LLM agents, how many
manipulate the agent's history or state mid-run and continue the agent from there?

\paragraph{Population.} The frame is the 200 index cards read for this project: a
convenience sample, not a systematic search.
Two independent LLM agents coded it, the second blind to the first.
Both read only our own secondhand summaries, no PDFs, so the codes describe the cards, not
the papers, and can err in both directions (sparse cards create false negatives, misreadings false
positives); silence was coded as silence.

\paragraph{Rubric and Stage 1.} Table~\ref{tab:census-rubric} states the frozen rubric.
Of the 200 cards, 41 passed Stage 1.
Both Stage 2 tables cover the same 41: the 41st (2603.24755) was promoted out of the excluded
set mid-census, and coder 2---given 40 cards first---had independently screened it in.

\begin{table}[h]
\centering
\small
\caption{The rubric as frozen before coding. Q3 is asked only when Q2 is a history-or-state
manipulation, and admits multiple codes.}
\label{tab:census-rubric}
\setlength{\tabcolsep}{4pt}
\begin{tabular}{@{}lp{9.6cm}@{}}
\toprule
 & Code set \\
\midrule
Stage 1 & screen from title, TL;DR and tags; include if the paper (a) studies LLM agents or multi-turn LLM systems \emph{and} (b) contributes primarily a taxonomy of failure modes, a failure-attribution or failure-localization method or benchmark, or an empirical diagnosis of why agents fail. Exclude capability benchmarks without failure analysis, methods papers whose failure discussion is incidental, surveys that do not themselves categorize failures, and position papers with no coded failure data \\
\midrule
Q1 label source & human annotation; LLM judge; rule-or-automatic; mixed; none; not stated in card \\
Q2 intervention & none; system-or-task change scored end-to-end; history-or-state manipulation at an intermediate point with continuation; other \\
Q3 class & healed history; provenance swap; placebo or length-matched synthetic history; counterfactual injection; environment-or-state reset; replay from checkpoint without content change; other \\
Q4 frozen state & does the paper restore an executable intermediate state (environment plus process) and continue the same agent from it? yes; prefix re-execution only; no; not determinable from card \\
Q5 evidence & one verbatim quote ($\le 30$ words) from the card supporting Q2 and Q4 \\
Q6 confidence & high; medium; low (low whenever the card is silent on a coded dimension) \\
\bottomrule
\end{tabular}
\end{table}

\begin{table}[h]
\centering
\small
\caption{Stage 2 counts, the two coders side by side ($N = 41$ each). Q2 and Q4 each
partition the 41 cards.}
\label{tab:census-counts}
\begin{tabular}{@{}lrr@{}}
\toprule
 & Coder 1 & Coder 2 \\
\midrule
Q2 = history-or-state manipulation with continuation & 22 & 13 \\
Q2 = system-or-task change scored end-to-end & 13 & 22 \\
Q2 = none & 6 & 6 \\
\midrule
Q4 = yes (frozen executable state, same agent continued) & 1 & 3 \\
Q4 = prefix re-execution only & 1 & 1 \\
Q4 = not determinable from card & 3 & 2 \\
Q4 = no & 36 & 35 \\
\bottomrule
\end{tabular}
\end{table}

\paragraph{Agreement.} Table~\ref{tab:census-counts} gives both coders' counts and Tables~\ref{tab:census-q2} and~\ref{tab:census-q4} the confusion matrices. On Q2 the coders agreed on 32 of 41 cards---78.0\%, Cohen's
$\kappa = 0.66$.
That is low, and one-sided: all nine disagreements are coder 1 coding a history-or-state
manipulation where coder 2 coded a system-or-task change scored end-to-end.
Only coder 2 wrote that boundary down beforehand, treating a run scored from its start
under a fixed condition as end-to-end.
On Q4 they agreed on 38 of 41---92.7\%, $\kappa = 0.70$.
Re-screening every fifth card of the excluded list alphabetically (32 of a 160-line list drawn
before the 41st card was promoted out of it), coder 2 agreed with coder 1's original screen on 31 of 32 and with the final revision on all 32; the one card that moved, 2603.24755, is the promotion described above.

\begin{table}[h]
\centering
\small
\begin{minipage}[t]{0.48\textwidth}
\centering
\caption{Q2 confusion (rows: coder 1).}
\label{tab:census-q2}
\setlength{\tabcolsep}{4pt}
\begin{tabular}{@{}lrrrr@{}}
\toprule
 & \multicolumn{3}{c}{Coder 2} & \\
\cmidrule(lr){2-4}
Coder 1 & man. & e2e & none & tot. \\
\midrule
manipulation & 13 & 9 & 0 & 22 \\
end-to-end & 0 & 13 & 0 & 13 \\
none & 0 & 0 & 6 & 6 \\
\midrule
total & 13 & 22 & 6 & 41 \\
\bottomrule
\end{tabular}
\end{minipage}\hfill
\begin{minipage}[t]{0.48\textwidth}
\centering
\caption{Q4 confusion (rows: coder 1).}
\label{tab:census-q4}
\setlength{\tabcolsep}{3pt}
\begin{tabular}{@{}lrrrrr@{}}
\toprule
 & \multicolumn{4}{c}{Coder 2} & \\
\cmidrule(lr){2-5}
Coder 1 & yes & pre. & no & n.d. & tot. \\
\midrule
yes & 1 & 0 & 0 & 0 & 1 \\
prefix only & 0 & 1 & 0 & 0 & 1 \\
no & 2 & 0 & 34 & 0 & 36 \\
not determ. & 0 & 0 & 1 & 2 & 3 \\
\midrule
total & 3 & 1 & 35 & 2 & 41 \\
\bottomrule
\end{tabular}
\end{minipage}
\end{table}

\paragraph{Frozen state, and the three Q4 disagreements.} Both coders coded 2608.06503
\citep{min2026trace} yes and Who\&When Pro, 2607.09996 \citep{liu2026whowhenpro}, prefix
re-execution only; coder 2 coded two more yes. Q4 counts throughout are the coders' assigned codes, not verified restorations. By our own reading of its protocol (Appendix~\ref{app:trace}), 2608.06503 re-executes a recorded prefix under a recipe hash, which coder 2's written convention would code as prefix re-execution only; we leave both coders' codes as recorded; harmonizing that one known inconsistency gives frozen-state counts of 0 and 2 and prefix-re-execution counts of 2 and 2, and the body reports the positively identified counts under both readings.
On ChainSWE, 2607.02606 \citep{jin2026chainswe}, both quoted the card's line that oracle
mode resets the container to the base commit and applies earlier gold patches before each
stage, which coder 2's convention makes a yes.
On 2608.13417 coder 2 called its own yes ``the least secure of the three,'' the card never
saying the running environment and process are restored.
Coder 1 quoted the same lines in both cases and recorded no reason for its no.
The third disagreement, 2607.24167, is not determinable for coder 1---local resume or
safe-state backtracking, environment and process unstated---and no for coder 2.

\paragraph{Not determinable.} Only two cards are not determinable for both coders: 2509.25370, which re-rolls out from the critical
step, and 2608.02464, which rolls a run back and re-runs it live.
Neither says whether state is restored or the prefix re-executed; coder 1 also marked
2607.24167.
They are the likeliest true frozen-state replays, so the coders positively identified one and three such papers (zero and two after harmonizing Min et al.), plus one or two prefix re-executions; because the cards are secondhand, no upper bound on the literature follows.

\paragraph{Limitations.} Two executing methods published outside the frozen frame, DoVer~\citep{ma2025dover} (checkpointed multi-agent debugging) and CausalFlow~\citep{bonagiri2026causalflow} (step replacement with downstream re-execution where an executor exists), would code as history-or-state manipulations with continuation; neither reports a replay-fidelity measurement, so neither changes the frozen-state counts.
The second coder recommended resolving the two shared not-determinable cards from the papers before publication. One (AgentDebug, arXiv 2509.25370) was read from the primary PDF after the census closed: it re-rolls out from the identified critical step with feedback, up to five attempts, but does not state how that step's state is reached and reports no fidelity measure, so its frozen-state code stays not determinable and no count changes; the other was not read, and the counts above are card-level.
One code per coder, no adjudication, no third rater: each $\kappa$
rests on one 41-card pairing, and the Q4 value sits on a table whose 34 no--no cells
dominate.
The inclusion check covers a fifth of the excluded cards. The paper reports the range.

\begin{table}[htbp]
\centering
\footnotesize
\caption{The 41 included cards, part 1 of 2 (cards 1--21). Q2: \emph{man.} =
history-or-state manipulation with continuation, \emph{e2e} = system-or-task change scored
end-to-end. Q4: \emph{pre.} = prefix re-execution only, \emph{n.d.} = not determinable from
card. The last column shows coder 2's code wherever it differs. Short titles are abridged from
each card's frontmatter.}
\label{tab:census-cards-a}
\setlength{\tabcolsep}{4pt}
\begin{tabular}{@{}rlp{4.4cm}lll@{}}
\toprule
\# & arXiv & Short title & Q2 & Q4 & Coder 2 differs \\
\midrule
1 & 2608.06503 & Toward Reliable Context Compression & man. & yes &---\\
2 & 2607.24167 & Falsifiable Commitment Planning & man. & n.d. & Q2 e2e; Q4 no \\
3 & 2601.23045 & The Hot Mess of AI & e2e & no &---\\
4 & 2509.09677 & The Illusion of Diminishing Returns & man. & no &---\\
5 & 2608.11772 & Diagnosis Before Recovery & e2e & no &---\\
6 & 2607.02606 & ChainSWE: multi-bug maintenance & man. & no & Q4 yes \\
7 & 2601.00828 & Decomposing LLM Self-Correction & man. & no & Q2 e2e \\
8 & 2607.22585 & The Scaffold Effect in Coding Agents & e2e & no &---\\
9 & 2608.15242 & LongRCA Bench & none & no &---\\
10 & 2606.05806 & When Tools Fail & man. & no &---\\
11 & 2608.19652 & Can Agent Memory Systems Track Evolving State? & man. & no & Q2 e2e \\
12 & 2608.13417 & Beyond Final Scores & man. & no & Q4 yes \\
13 & 2604.02547 & Beyond Resolution Rates & none & no &---\\
14 & 2606.29718 & Diagnosing and Mitigating Context Rot & man. & no & Q2 e2e \\
15 & 2607.07989 & Who Broke the System? & none & no &---\\
16 & 2605.24279 & ContextEcho & man. & no &---\\
17 & 2605.27898 & A Unified Framework for Agentic Evaluation & e2e & no &---\\
18 & 2607.09996 & Who\&When Pro & man. & pre. &---\\
19 & 2603.04474 & From Spark to Fire & man. & no & Q2 e2e \\
20 & 2605.24197 & A Sober Look at Agentic Misalignment & e2e & no &---\\
21 & 2602.07338 & Intent Mismatch Causes LLMs to Get Lost & man. & no & Q2 e2e \\
\bottomrule
\end{tabular}
\end{table}

\begin{table}[htbp]
\centering
\footnotesize
\caption{The 41 included cards, part 2 of 2 (cards 22--41). Columns as in
Table~\ref{tab:census-cards-a}.}
\label{tab:census-cards-b}
\setlength{\tabcolsep}{4pt}
\begin{tabular}{@{}rlp{4.4cm}lll@{}}
\toprule
\# & arXiv & Short title & Q2 & Q4 & Coder 2 differs \\
\midrule
22 & 2505.00212 & Which Agent Causes Task Failures? & none & no &---\\
23 & 2606.06324 & From Failed Trajectories to Reliable Agents & e2e & no &---\\
24 & 2505.23840 & Measuring Sycophancy in Multi-turn Dialogues & e2e & no &---\\
25 & 2606.22528 & Governance Decay & man. & no &---\\
26 & 2607.09510 & Failure as a Process & none & no &---\\
27 & 2603.24631 & Coherence Collapse & e2e & no &---\\
28 & 2603.06870 & LEAD: the no-recovery bottleneck & man. & no & Q2 e2e \\
29 & 2503.13657 & Why Do Multi-Agent LLM Systems Fail? & e2e & no &---\\
30 & 2608.18066 & On the Fragility of Self-Improving Agents & e2e & no &---\\
31 & 2310.01798 & LLMs Cannot Self-Correct Reasoning Yet & man. & no & Q2 e2e \\
32 & 2606.07889 & Strained Coherence & none & no &---\\
33 & 2605.08060 & The Memory Curse & man. & no &---\\
34 & 2509.25370 & Where LLM Agents Fail and How They Learn & man. & n.d. &---\\
35 & 2608.02464 & Real-Time Detection and Repair of Failures & man. & n.d. &---\\
36 & 2505.06120 & LLMs Get Lost in Multi-Turn Conversation & e2e & no &---\\
37 & 2604.11978 & The Long-Horizon Task Mirage? & e2e & no &---\\
38 & 2505.02709 & Evaluating Goal Drift in LM Agents & man. & no &---\\
39 & 2502.15840 & Vending-Bench & e2e & no &---\\
40 & 2502.19559 & Stay Focused: problem drift in debate & man. & no &---\\
41 & 2603.24755 & SlopCodeBench & man. & no & Q2 e2e \\
\bottomrule
\end{tabular}
\end{table}
}{}

\clearpage
\section{Use of large language models}\label{app:llm}

LLM agents were used as instruments in two reported measurements, as stated where they occur: the same-agent re-labeling pass of Measurement~3.1 (Appendix~\ref{app:carrier}) and the two-coder literature census (Appendix~\ref{app:census}). LLM assistance was additionally used in the engineering of the system and benchmark code and in manuscript preparation, under the authors' direction and review; the authors take full responsibility for all content, claims, and measurements.

\end{document}